\documentclass[letterpaper, 10 pt, conference]{ieeeconf}  % Comment this line out if you need a4paper

\IEEEoverridecommandlockouts                              % This command is only needed if 
\usepackage{multicol}
\usepackage[bookmarks=true]{hyperref}
\usepackage{multirow}
\usepackage{times}
\usepackage{epsfig}
\usepackage{graphicx}
\usepackage{amsmath}
\usepackage{amssymb}
\makeatletter
\let\NAT@parse\undefined
\makeatother
\usepackage[numbers]{natbib}

\usepackage{multicol}
\usepackage{graphicx}
\usepackage{booktabs}
\usepackage[table]{xcolor}
\usepackage{tabularx}
\usepackage[font=small,labelfont=bf]{caption}

\renewcommand{\baselinestretch}{0.935}

\title{\LARGE \bf Visuomotor Control in Multi-Object Scenes Using Object-Aware Representations}

\author{Negin Heravi*, Ayzaan Wahid, Corey Lynch, Pete Florence, Travis Armstrong, \\ Jonathan Tompson,
Pierre Sermanet, Jeannette Bohg, Debidatta Dwibedi
\thanks{
* Work done as an intern at Google. 
 N.\ Heravi was also with Stanford University during this work. A.\ Wahid, C.\ Lynch, P.\ Florence, T.\ Armstrong, J.\ Tompson, P.\ Sermanet, and D.\ Dwibedi are with Robotics at Google.  J.\ Bohg is with Stanford University. {\tt\small nheravi@alumni.stanford.edu}. 
\newline
Toyota Research Institute ("TRI") provided funds to assist the author with their research, but this article solely reflects the opinions and conclusions of its authors and not TRI or any other Toyota entity.
}
}

\begin{document}

\maketitle              % typeset the title of the contribution

\thispagestyle{empty}
\pagestyle{empty}

\begin{abstract}
Perceptual understanding of the scene and the relationship between its different components is important for successful completion of robotic tasks. Representation learning has been shown to be a powerful technique for this, but most of the current methodologies learn task specific representations that do not necessarily transfer well to other tasks. Furthermore, representations learned by supervised methods require large, labeled datasets for each task that are expensive to collect in the real-world. Using self-supervised learning to obtain representations from unlabeled data can mitigate this problem. However, current self-supervised representation learning methods are mostly object agnostic, and we demonstrate that the resulting representations are insufficient for general purpose robotics tasks as they fail to capture the complexity of scenes with many components. In this paper, we show the effectiveness of using object-aware representation learning techniques for robotic tasks. Our self-supervised representations are learned by observing the agent freely interacting with different parts of the environment and are queried in two different settings: (i) policy learning and (ii) object location prediction. We show that our model learns control policies in a sample-efficient manner and outperforms state-of-the-art object agnostic techniques as well as methods trained on raw RGB images. Our results show a 20\% increase in performance in low data regimes (1000 trajectories) in policy training using implicit behavioral cloning (IBC). Furthermore, our method outperforms the baselines for the task of object localization in multi-object scenes. Further qualitative results are available at \href{https://sites.google.com/view/slots4robots}{https://sites.google.com/view/slots4robots}.
% We would like to encourage you to list your keywords within
% the abstract section using the \keywords{...} command.
% \keywords{ Robot Learning, Robot Vision, AI-enabled Robotics}
% ICRA doesn't have keywords inside the paper
\end{abstract}
\begin{figure}[h!]
    \noindent
    \centering
    \includegraphics[width=0.9\linewidth]{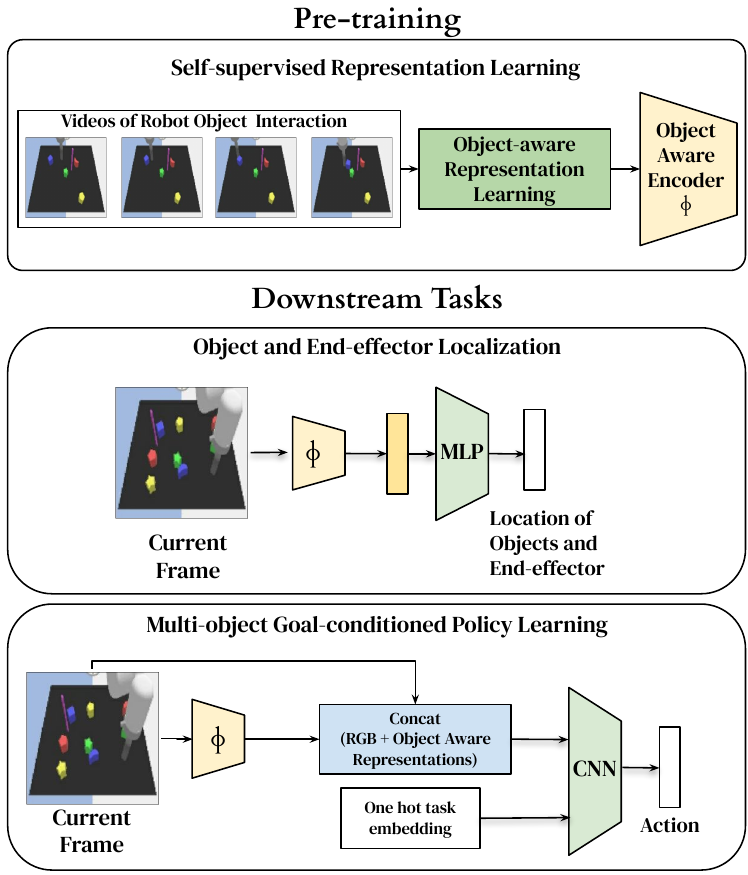}
    \caption{We train a perception encoder using object-aware self-supervised learning. Then, we freeze the encoder weights and demonstrate this encoder is suitable for downstream tasks such as object localization and multi-object goal-conditioned policy learning.}
    \label{fig:vision_figure}
    \vspace{-24pt}
\end{figure}

\section{Introduction}

General purpose robots need to encode information about their environment and themselves in a way that is not task specific and can be easily transferred to new situations and tasks. Current techniques for learning representations of the robot's environment are often trained in a supervised manner on task-specific datasets. However, collecting and labeling a new dataset per task is not scalable. Self-supervised learning methods that aim to learn representations from unlabeled data hold the potential to help robots learn about their environments without manual annotation.

Prior work in robotics has shown that performance and sample-efficiency of policy learning improves with self-supervised scene representations. These methods use compact representations in form of a global embedding~\cite{Sermanet2017TCN}, sparse key-point coordinates~\cite{finn2016deep, kulkarni2019unsupervised, florence2019self, manuelli2020keypoints}, or other object embeddings~\cite{yuan2021sornet} as input to policy learning modules. In this work, we explore a class of self-supervised models called Slot Attention~\cite{locatello2020objectcentric} for representation learning in a robotic setup and differ from these prior works by learning the policy directly on dense per-pixel features and object masks produced by the Slot Attention model. This is particularly important in the context of multi-object manipulation as the representations need to encode the location of multiple objects in the scene along with the end-effector. Additionally, learning slot based representations does not require multi-view cameras~\cite{Sermanet2017TCN,florence2019self,manuelli2020keypoints} or canonical images of objects~\cite{kulkarni2019unsupervised,yuan2021sornet}.

Slot Attention models use sequential attention based mechanisms to group low level features in a scene where each group falls into a slot bin \cite{locatello2020objectcentric}. The authors show that the model can segment objects in an unsupervised manner. Inspired by this architecture, we propose to use these models, that can learn the extent of multiple objects in a scene, to learn representations that can be used for a variety of downstream robotic tasks. Our hypothesis is that the abstracted information using Slot Attention can improve data sample efficiency and performance in downstream training since it is object-aware making it suitable for extracting information in multi-object scenes. We test this hypothesis in the tasks of object localization and multi-object goal-conditioned policy learning discussed in detail in the next sections. Our model is trained in multiple stages. First, we train Slot Attention in the object discovery mode in a self-supervised manner. Then, we freeze the weights and use these learned representations to train different small downstream networks for each task. We train our models using data of a robot interacting with blocks of different shapes and colors placed on a table in a simulation environment. Using this setup, we study the gain in performance by using these representations in different data regimes. We observe that the mask and features learned by our model boost performance on both tasks of object localization and behavioral cloning. Particularly in the low data regime, our features result in a 20\% improvement in task completion success rate.

In summary, the contributions of our work are as follows: 
\begin{enumerate}
     \item We show that our Slot Attention inspired representations encode location and properties of all objects in the scene while object agnostic self-supervised methods such as MoCo \cite{he2019moco} only focus on a few objects.  
    \item We show that our method needs fewer supervised action labels to learn policies (i.e. it is more sample efficient) and learns policies that have faster training convergence than alternative state-of-the-art methods.
    % \item As a proxy for real robot settings, we demonstrate our method's capability for prediction the action that happened between time separated frames of a robot action video. Our learnt representations are shown to be effective in improving prediction performance in this regime. We hypothesize that representations that are successful in this task will transfer well to real world policy learning.
    
\end{enumerate}

\section{Related Work}

\textbf{Self-supervised Representation Learning.} Self-supervised learning has been immensely successful in training large models without any labels across different visual modalities: images~\cite{chen2020simple,he2019moco}, optical flow~\cite{jonschkowski2020matters} and videos~\cite{wang2015unsupervised, han2020coclr}. As these methods do not rely on manual labeling, they are well-suited for learning features from videos of robots interacting with objects. Self-supervised losses tend to be either a contrastive or a reconstruction loss. In this work, we compare features learned with these two different losses as input to a policy-learning network. In particular, we focus on a class of models that use the reconstruction loss but also the notion of slots/objects that induces grouping of similar pixels into slots in a self-supervised way \cite{locatello2020objectcentric}. 

\textbf{Self-supervised Learning for Robotics.} Several previous papers have explored self-supervised learning methods for robotics. \cite{finn2016deep} learned features as input to policies using a spatial-softmax bottleneck layer and a reconstruction loss. ~\cite{Sermanet2017TCN, dwibedi2018learning} proposed a self-supervised approach for learning embeddings based on metric learning using videos from multiple cameras. These embeddings were shown to be useful in reward calculation or as input for reinforcement learning methods.
\cite{jang2018grasp2vec} also used self-supervised embeddings to improve grasping. \cite{lee2019making} used self-supervised learning of multimodal representations for contact-rich manipulation tasks. \cite{deng2020self} proposed a self-supervised approach to improve 6D pose estimation of objects by using a robot that interacts with objects. \cite{zakka2021xirl} explored how features from self-supervised learning methods can be used as a reward for training unseen reinforcement learning agents.

\textbf{Object-centric Representations for Robotics.}
% Similar to prior work~\cite{jang2018grasp2vec,lin2020learning, wang2019deep, devin2018deep}, we also explore if the addition of object-centric information can improve policy learning.
Various types of object-centric representations have been explored for use in policies in robotics, and although sometimes these use supervised learning~\cite{devin2018deep,wang2019deep,manuelli2019kpam}, often these may be self-supervised~\cite{jang2018grasp2vec} as well.
% But unlike these papers that focus on improving downstream policy learning by using features or outputs of models trained with supervised learning, we use features from models trained with self-supervised losses.
In this aspect, our work is similar in spirit to works which use self-supervision to acquire embeddings interpreted as keypoints either learned through autoencoding bottlenecks ~\cite{finn2016deep,kulkarni2019unsupervised}, or multi-view consistency ~\cite{florence2018dense, florence2019self, manuelli2020keypoints}. In contrast to these papers however, we directly use dense features (the same resolution as the input image) rather than sparsifying the representation down to keypoints.

Another closely related work is APEX~\cite{wu2021apex} which shows how segmentation masks can improve performance when using a heuristic policy. We differ from their work in using the Slot Attention algorithm for representation learning and directly using the learned masks for policy learning. We find that these features are crucial for task success when the number of objects in the scene increases.
%We study the utility of Slot Attention mask instead of segmentation masks for end-to-end behavioral cloning without

\textbf{Imitation Learning.} Past work~\cite{florence2019self} has explored the use of self-supervised features to improve the performance of behavior cloning models. In this work, we also investigate the effectiveness of self-supervised features but in the context of a modern behavior cloning algorithm, Implicit Behavior Cloning (IBC)~\cite{florence2021implicit}, which was shown to be better than standard behavior cloning on a number of robotic tasks.

% \textbf{Multi-task Policy Learning.}
% \cite{lynch2019play, shao2020concept}

% Slot Attention showed an object-centric representation learning technique that learns to segment objects in a scene in a self-supervised manner. However, the paper only shows the performance on synthetic datasets with simple objects and shapes such as CLEVR. 
% Yang et al took inspiration from the Slot attention algorithm for video object segmentation by motion grouping. Their algorithm uses optical flow to segment a moving object in complex real world scenarios. However, they are only able to segment a scene into foreground and background (only segment the object of interest) while we are looking into segmenting multiple objects in a scene in our work also flow calculation is expensive can't be done in real time.

% SAVI extends the Slot Attention algorithm for videos but requires explicit conditioning of the center of mass of the objects or the bounding box in the first frame.

\section{Approach}
\label{approach:slot}
Our overall framework learns object aware representations from unlabeled videos using Slot Attention~\cite{locatello2020objectcentric}. We then freeze the weights of the representation architecture, and use features from this model for downstream robotic tasks of object and end-effector localization (Section \ref{task:locationpred}) as well as policy learning (Section \ref{task:policylearning}). We show an overview of our approach in Figure \ref{fig:vision_figure}. 

%With this setting, we show that our representation network has encoded the observations in a task agnostic generally purposeful setting making it useful for diverse scenarios. Commenting <Debi:Some reviewers will find this overclaiminig>

\begin{figure}[h!]
    \noindent
    \centering
    \includegraphics[width=0.9\linewidth]{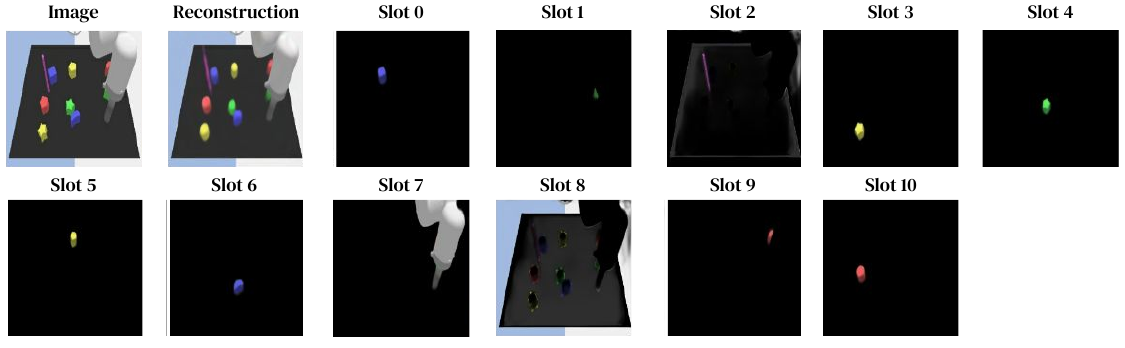}
    \caption{\textbf{Qualitative example of masks learned by Slot Attention.} Slot Attention is able to localize objects and the end-effector by observing interaction videos without any mask-level supervision.}
    \label{fig:slot_qualitative}
    \vspace{-8pt}
\end{figure}

%\subsection{Representation Learning Method}
%\label{approach:slot}
As our representation method, we use an image encoder $\phi$ that takes an RGB image $I$ as input, and outputs an embedded feature representation $\phi(I)$. Given an input batch of N images $\{ I_1, ..., I_N\}$, we train $\phi$ using a variation of the Slot Attention network~\cite{locatello2020objectcentric} which consists of a convolution-based encoding architecture $\phi$ followed by an iterative attention mechanism \cite{vaswani2017attention}. This architecture is designed such that it groups the features of an image $I_i$ into $K$ slots where $K$ is a hyperparameter. To achieve this, it uses an iterative dot-product based attention mechanism \cite{vaswani2017attention} that binds image features (input keys) and slots (queries). The attention coefficients are normalized over the $K$ slots using a Softmax operation and are used to update the slots in each iteration using a {\em Gated Recurrent Unit\/} (GRU). This normalization over slots makes them compete with each other and each specialize in explaining a different component in the image. 
Given these slots, an upsampling convolutional decoder spatially broadcasts and reconstructs each slot separately into image slots $R_{ik}$ as well as the corresponding spatial alpha masks $M_{ik}$. The masks are then used as weights to sum the reconstruction slots into a single combined reconstructed image for input $I_i$. This results in self-supervised decomposition of low-level image features into abstract groups. We train this network in a self-supervised manner using the L2 pixel-wise difference of the reconstructed and the original image: 
\begin{equation}
 L = \frac{\sum_{i=1}^{N}(I_i-\sum_{k=1}^{K}M_{ik} \times R_{ik})^2}{N}   
\end{equation}

We modified the Slot Attention architecture in two ways. First, like \cite{yang2021self}, we initialize the slots to be learnable fixed vectors instead of samples from a learned Gaussian distribution. This change mitigated slot swapping between objects as the noise in the Gaussian setting could lead to permutations. Second, we use convolutions followed by upsampling instead of transposed convolutions to prevent checkerboard artifacts~\cite{odena2016deconvolution}. Please refer to \cite{locatello2020objectcentric} for more training details of the Slot Attention algorithm. Figure \ref{fig:slot_qualitative} shows qualitative examples of the performance of training this model on our robotic dataset. 
 
After training this network, we freeze the weights and use the output of the convolution-based encoder $\phi(I)$ as representation in our policy learning.  The pre-trained frozen Slot Attention models used in our experiments were trained for $500k$ steps with a batch size of $8$ on images of size $160$ by $320$ and $K=16$ slots with random seed initialization unless otherwise noted. For the experiments where the fraction of the data available varies, the slot representations are also only trained on the selected fraction of data. For the localization task, we use the dense masks ($M_{1},..,M_{K}$) as input to the downstream network. These masks encode the location of objects in pixel space in the form of slots. 

\section{Experiments and Discussion}
\label{experiments}
The goal of our experiments is to evaluate whether self-supervised object aware representations learned using Slot Attention provide performance gain for robotic tasks. To quantify this, we compare various representation learning techniques for tasks of object localization, multi-object goal-conditioned policy learning, and action prediction. 

\subsection{Baselines}
We compare our method with the following baselines:

\noindent\textbf{MoCo~\cite{he2019moco}}. We train an encoder using a contrastive loss in a self-supervised manner as outlined in \cite{chen2020improved}. The output of the encoder is spatially averaged to produce an embedding. MoCo uses this contrastive loss to learn invariances of an image going through various data augmentations. In the context of this loss, positives are provided by using the embedding from a momentum encoder with the augmented version of the same image as input and negatives are sampled from a queue that keeps past embeddings in memory. We use this method as we want to compare the performance of contrastive losses and reconstruction losses (like that used in Slot Attention) for robotic tasks. We use an embedding size of 128, queue size of 16384, softmax temperature of 0.1 and batch size of 16 to train the MoCo encoder.

\noindent\textbf{Autoencoder}. We train an encoder-decoder architecture using the reconstruction loss~\cite{pmlr-v27-baldi12a}. This method has the same loss as the slot-encoder architecture, but it does not use any notion of slots/objects. The objective of this baseline comparison is isolating the importance of the Slot Attention module and the reconstruction loss for learning representations.

\subsection{Environment} 
\label{environment} We use a robotic environment implemented in PyBullet~\cite{coumans2016pybullet}. In our setup, a robot arm is attached to a fixed base such that it can manipulate objects in front of it on a table. A cylinder is attached to the end of the arm which serves as the end-effector to push around blocks of different shapes and colors on a 2D plane (similar to \cite{coreylynch}). We use this environment for collecting the data for all the simulation experiments in the following sections.

\subsection{Object Localization: Task and Metrics}
\label{task:locationpred}
In this experiment, we investigate if the Slot Attention architecture can encode information about \textit{all} the objects present in a scene. This property is important for learning policies on datasets with multiple objects and for tasks that depend on full scene information such as object localization. We evaluate the representations on the object localization task in a simulated environment which provides ground truth object locations. We only use this ground truth during downstream task training not for representation learning. 
% Representation learning methods that tend to only focus on a few objects do not provide significant gains on multi-object manipulation.

We learn representations from a dataset of demonstrations collected in simulation environment described in Section \ref{environment}. Then, we freeze the weights of the representation network, and train an MLP to predict the location of the center of each block in robot coordinates. Please note that these locations are predicted in the robot coordinate base as ground truth is available in this space and not in image space. As a result, the network is learning hand-eye-coordination on top of finding the location of each block in the image. The MLP used for this task consists of two fully connected layers of size 256 and outputs the 2D location of all the blocks and the end-effector on the table's plane. For Slot Attention in this task, the input to our downstream MLP is the center of mass of the predicted slot masks. For our MoCo baseline, we use the output of the image encoder as input. For our Autoencoder baseline, as the MLP decoder needs an input vector, we use global average pooling layer on top of the output of the encoder CNN and use that as the input.

As an interpretable evaluation metric, we use {\em Probability of Correct Keypoint} (PCK)~\cite{yang2012articulated} which captures the percentage of times an object location is predicted correctly. This metric considers an object to be correctly localized if the predicted coordinates are within a given threshold of the ground truth. This is a commonly used method in computer vision research to evaluate localization of human and object keypoints \cite{yang2012articulated}. We chose a PCK threshold value of 0.1 of the length of the table (about 5 cm). 

We test the performance of our method with 1, 4, and 8 blocks. For the Slot Attention models, we use 7, 11, and 11 slots for the 1, 4, and 8 blocks respectively. The number of slots were chosen based on their performance on the reconstruction loss when pretraining the model independent of the downstream task. We train models with a dataset of 160k trajectories with image size of $256\times256$ for 150k steps each with batch size of 16 on one V100 GPU. We use the checkpoint with the lowest loss during representation learning for downstream training of object localization. 

% The motivation for using a smaller sized image in these experiments was to be able to directly compare the two methods together using larger batch sizes as MoCo does not work well with small batch sizes while Slot Attention runs out of memory with larger batchsize for the original 160x320 image size. These experiments were run on 1 V100 GPU.

\subsection{Object Localization: Results}
\begin{figure}
    \noindent
    \centering
    \includegraphics[width=0.72\linewidth]{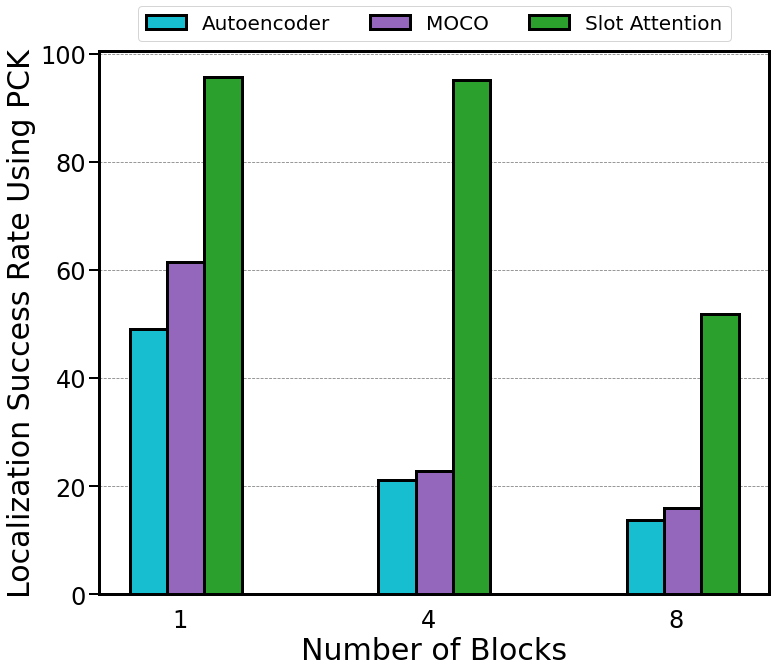}
    \caption{Performance comparison on object localization over number of blocks present in the scene using {\em Probability of Correct Keypoint} (PCK) as evaluation metric. Higher percentage indicates better performance. The baselines are able to learn the location of one object in the scene (the end-effector) resulting in a low performance on average that decreases as the number of blocks increases. Slot Attention is able to localize multiple objects but sometimes struggles with objects of same color but different shapes in the 8 block scenario.}
    \label{fig:obj_decode_image_blocks}
    \vspace{-10pt}
\end{figure}

Slot Attention outperformed the baselines in all the object localization experiments specially in multi-object cases as seen in Tables \ref{tab:obj_decode_1block}-\ref{tab:obj_decode_8blocks} and Figure \ref{fig:obj_decode_image_blocks}. We further made the following observations:

\noindent \textbf{Baselines struggle with localizing multiple objects.}
Table \ref{tab:obj_decode_1block} and \ref{tab:obj_decode_4blocks} show the performance of our method on the cases of 1 and 4 blocks respectively, and Table \ref{tab:obj_decode_8blocks} shows the performance on the 8 block scenario. We observe that while MoCo and Autoencoder are able to learn to predict object location more accurately on a dataset with a single object, they struggle to encode object locations in multi-object scenes. In Figure~\ref{fig:obj_decode_image_blocks}, we also show how the average object localization performance of different representations varies as the number of objects in the scene changes. 
%Figure \ref{fig:obj_decode_image} shows a qualitative example of the performance of these models on an example evaluation image. 

% \noindent \textbf{Slot Attention is able to localize the target rod}
% We also observe that Slot Attention model localizes thin objects like the purple rod in the scene. This is important for the policy learning method to know where the target location is.

\noindent \textbf{Slot Attention struggles with localizing objects with similar color and shape.}
While outperforming the baselines, we observe that the Slot Attention model finds it difficult to localize blocks of the same color but with fine-grained differences in their shape. In Table~\ref{tab:obj_decode_8blocks}, we show the Slot Attention model gives poor localization performance for the two yellow and the two green objects that have similar shapes due to slot assignment swapping between similar looking blocks. This is due to Slot Attention being trained using a pixel-wise image reconstruction loss, and hence, struggling to differentiate subtle differences in shapes when the color is the same. Subtle shape differences only contribute to a small number of pixel differences resulting in slight changes in the loss. However, Slot Attention still outperforms the other methods for this task. 

% For the 8 block scenario, slot sometimes gets confused with objects of the same color resulting in swapping of the slots and the decreased performance observed in the 8 block case scenario. However, even in the 8 block scenario Slot Attention significantly outperforms the MoCo baseline. Furthermore, it is possible that the slot swapping is an artifact of the position embedding decoder that generate the masks used in this experiment while the slot feature representation does not suffer from a similar problem. Even in cases where the slot information are swapping in the feature space the information encoded in the representation can be helpful for other downstream tasks where knowing the occupied locations in the environment is important but knowledge of the exact location of each object is not critical for task performance.

% \begin{figure}[h!]
%     \noindent
%     \centering
%     \includegraphics[width=0.9\linewidth]{Figs/Figure2-objectdecode.pdf}
%     \caption{Qualitative example of the model's performance on object location prediction. As baseline, performance of a representation learned using MoCo is shown on the middle column. In a perfect prediction, shapes should overlap with the corresponding matching circles of those objects. It can be seen that this is the case for Slot Attention in the 1 and 4 block case. For the 8 block case, it can be seen that the predicted slots are closer to the ground truth predictions than that of the MoCo baseline.}
%     \label{fig:obj_decode_image}
% \end{figure}

\begin{table}[]
\centering
\caption{Performance of different self-supervised representations on downstream task of object location prediction (1 block case) using {\em Probability of Correct Keypoint} (PCK) as metric. Higher percentage indicates better performance.}
\begin{tabularx}{\linewidth}{l|c|c|c}
&   \multicolumn{3}{c}{\textbf{Object}}\\
\midrule
\textbf{Input to object localizer} & 
\multicolumn{1}{c|}{\textbf{Mean}} &   \multicolumn{1}{c|}{\textbf{End Effector}} &       \multicolumn{1}{c}{\textbf{Blue Cube}} \\
\midrule
% \textbf{Input to object localizer ($\downarrow$)} &     \multicolumn{2}{c} {\textbf{PCK@0.1}}\\
% \toprule
MoCo    &61.5
&83.1
&39.9

           \\
\midrule
Autoencoder   &49.0
&82.5
&15.5

 \\
\midrule
 Slot Attention    
&\textbf{95.8}
&\textbf{97.2}
&\textbf{94.4}

                 \\  

\end{tabularx}

\label{tab:obj_decode_1block}
\end{table}

\begin{table}
\centering
\caption{Performance of different self-supervised representations on the task of object location prediction (4 blocks case) using {\em Probability of Correct Keypoint} (PCK) as metric. Higher percentage indicates better performance.}
\begin{tabular}{l|c|c|c|c|c|c}
&   \multicolumn{6}{c}{\textbf{Object}}\\
\midrule
\textbf{Input to}  & \multicolumn{1}{c|}{\textbf{Mean}}  & \multicolumn{1}{c|}{\textbf{End }}  &       \multicolumn{1}{c|}{\textbf{Red}} &       \multicolumn{1}{c|}{\textbf{Blue}} &       \multicolumn{1}{c|}{\textbf{Green}}&       \multicolumn{1}{c}{\textbf{Yellow}} \\
\textbf{object localizer}  & 
\multicolumn{1}{c|}{\textbf{}}  & \multicolumn{1}{c|}{\textbf{Eff.}}  &       \multicolumn{1}{c|}{\textbf{Moon}} &       \multicolumn{1}{c|}{\textbf{Cube}} &       \multicolumn{1}{c|}{\textbf{Star}}&       \multicolumn{1}{c}{\textbf{Pent.}} \\
\midrule

% \toprule
MoCo  &22.9
&66.3
&15.2
&14.7
&10.3
&8.2

  \\

\midrule
Autoencoder &21.2
&57.6
&12.4
&11.1
&11.8
&13.2

 \\
\midrule
 Slot Attention & \textbf{95.1}	& \textbf{94.7}	& \textbf{94.6}	& \textbf{96.4}	& \textbf{94.5}	& \textbf{95.2}

       \\  

\end{tabular}

\label{tab:obj_decode_4blocks}
\end{table}

\begin{table*}[h!]
\centering
\footnotesize
\caption{Performance of different self-supervised representations on downstream task of object location prediction (8 blocks case) using {\em Probability of Correct Keypoint} (PCK) as metric. Higher percentage indicates better performance.}
\begin{tabular}{l|c|c|c|c|c|c|c|c|c|c}
% \begin{tabular}{@{} p{0.8cm}|p{0.4cm}|p{0.4cm}|p{0.4cm}|p{0.4cm}|p{0.4cm}|p{0.4cm}|p{0.4cm}|p{0.4cm}|p{0.4cm}|p{0.4cm} @{}}
&   \multicolumn{10}{c}{\textbf{Object}}\\
\midrule
{\textbf{Input to}}
  & \multicolumn{1}{c|}{\textbf{Red}} &       \multicolumn{1}{c|}{\textbf{Blue}} &       \multicolumn{1}{c|}{\textbf{Green}}&       \multicolumn{1}{c|}{\textbf{Yellow}}&       \multicolumn{1}{c|}{\textbf{Red}}&       \multicolumn{1}{c|}{\textbf{Blue}}&       \multicolumn{1}{c|}{\textbf{Green}}&       \multicolumn{1}{c|}{\textbf{Yellow}}
  &       \multicolumn{1}{c|}{\textbf{End}}&       \multicolumn{1}{c}{\textbf{Mean}}\\
\textbf{object localizer} &         \multicolumn{1}{c|}{\textbf{Moon}} &       \multicolumn{1}{c|}{\textbf{Cube}} &       \multicolumn{1}{c|}{\textbf{Star}}&       \multicolumn{1}{c|}{\textbf{Pentagon}}&       \multicolumn{1}{c|}{\textbf{Pentagon}}&       \multicolumn{1}{c|}{\textbf{Moon}}&       \multicolumn{1}{c|}{\textbf{Cube}}&       \multicolumn{1}{c|}{\textbf{Star}}
  &       \multicolumn{1}{c|}{\textbf{Effector}}&       \multicolumn{1}{c}{\textbf{}}
\\
\midrule
%  &     \multicolumn{9}{c} {\textbf{PCK@0.1}}\\
% \toprule
MoCo 
&9.9
&13.7
&10.5
&7.4
&11.9
&13.1
&9.8
&9.1
&58.4
&16.0

\\
\midrule
Autoencoder 	&9.1	&14.4&	7.3&	9.9&	12.0	&13.7	&8.5	& 8.3 &41.0 &13.8
 \\

\midrule
 Slot Attention 
&\textbf{79.8}
&\textbf{79.4}
&\textbf{20.2}
&\textbf{18.4}
&\textbf{86.9}
&\textbf{83.0}
&\textbf{23.0}
&\textbf{15.6}
&\textbf{60.1}
&\textbf{51.8}

      \\  
% \midrule
% \textbf{} &     \multicolumn{9}{c} {\textbf{PCK@0.25}}\\
% \toprule
% MoCo    &  63.41&  93.4  &  60.99     & 63.60             & 59.03& 60.51& 65.58       & 55.43              &57.80   & 54.37     \\

% \midrule
% \rowcolor{green!50}Slot    &90.11 &   99.4            & 97.36      & 96.39 & 82.45 & 79.15&  98.52    & 98.58              & 80.19   & 78.97         \\  
\end{tabular}

\label{tab:obj_decode_8blocks}
\end{table*}
% \begin{table*}[h!]
% \centering
% \begin{tabular}{l|c|c}
% % \begin{tabular}{@{} p{0.8cm}|p{0.4cm}|p{0.4cm}|p{0.4cm}|p{0.4cm}|p{0.4cm}|p{0.4cm}|p{0.4cm}|p{0.4cm}|p{0.4cm}|p{0.4cm} @{}}
% &   \multicolumn{2}{c}{\textbf{Object (continued)}}\\
% \midrule
% {\textbf{Input to object localizer}}
%   &  \multicolumn{1}{c|}{\textbf{End Effector}} 
%   & \multicolumn{1}{c}{\textbf{Mean}}\\

% \midrule
% %  &     \multicolumn{9}{c} {\textbf{PCK@0.1}}\\
% % \toprule
% MoCo 
% &58.4
% &16.0

% \\
% \midrule
% Autoencoder &41.0 & 13.8		
%  \\

% \midrule
%  Slot Attention  &\textbf{60.1}
%  & \textbf{51.8}

%       \\  
% % \midrule
% % \textbf{} &     \multicolumn{9}{c} {\textbf{PCK@0.25}}\\
% % \toprule
% % MoCo    &  63.41&  93.4  &  60.99     & 63.60             & 59.03& 60.51& 65.58       & 55.43              &57.80   & 54.37     \\

% % \midrule
% % \rowcolor{green!50}Slot    &90.11 &   99.4            & 97.36      & 96.39 & 82.45 & 79.15&  98.52    & 98.58              & 80.19   & 78.97         \\  
% \end{tabular}
% \end{table*}

\begin{table*}[h!]
\centering
\caption{Performance comparison on policy learning using the rate of successful task completion as metric. Higher success rate indicates better performance. Bold values show the method with maximum performance without access to ground truth information. The method with access to ground truth segmentation provides an upper bound for this task (oracle).
}
\begin{tabular}{l|cc|cc|cc|cc}
\textbf{No. of training episodes ($\rightarrow$)} &  \multicolumn{2}{c|}{\textbf{1000}} &       \multicolumn{2}{c|}{\textbf{2000}} &   \multicolumn{2}{c|}{\textbf{3000}}  &   \multicolumn{2}{c}{\textbf{10000}}\\
\midrule
\textbf{Input to Policy ($\downarrow$)} & \textbf{Mean } & \textbf{Std. Dev.} & \textbf{Mean } & \textbf{Std. Dev.} & \textbf{Mean } & \textbf{Std. Dev.}& \textbf{Mean } & \textbf{Std. Dev.}\\
\midrule
RGB     &36.9
&5.8
&78.9
&8.4
&88.8
&3.2
&95.1
&2.8\\
RGB+Groundtruth Segmentation (oracle)  &78.5
&5.0
&93.5
&1.3
&94.8
&1.3
&95.4
&0.8\\ 
% \midrule
% RGB + MoCo       &35.9
% &9.2
% &77.1
% &3.1
% &84.9
% &6.7
% &92.2
% &2.3\\
\midrule
Autoencoder          &46.0
&3.2
&61.3
&9.8
&83.5
&3.8
&85.8
&6.1\\
RGB+Autoencoder      &49.0
&13.1
&76.9
&5.3
&66.5
&32.3
&92.6
&1.9\\ 
\midrule
Slot Attention   &\textbf{57.1}
&1.5
&86.0
&1.2
&\textbf{92.4}
&1.9
&\textbf{95.0}
&1.7\\  
RGB + Slot Attention  &53.3
&9.9
&\textbf{87.8}
&4.6
&92.6
&2.6
&95.0
&1.3\\
\end{tabular}
\label{tab:policy_learning}
\end{table*}
\begin{figure}
    \noindent
    \centering
    \includegraphics[width=\linewidth]{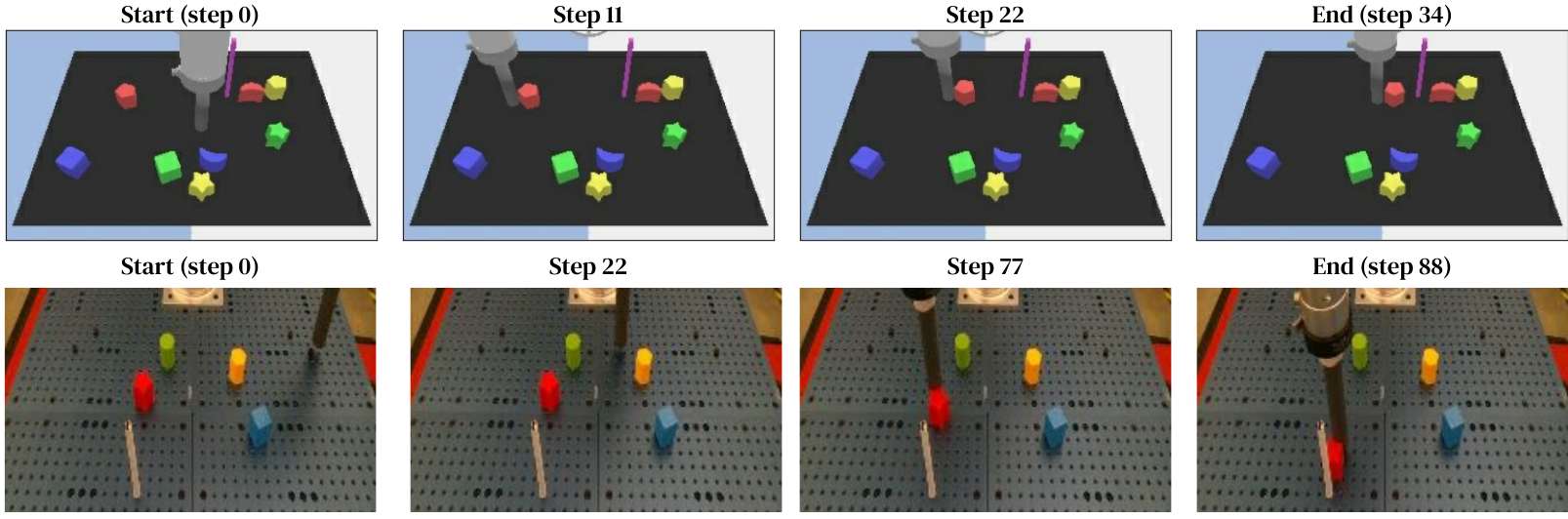}
    \caption{Example demonstrations: moving the red pentagon to pole in sim (top) and moving the red star to the pole in real (bottom).}
    \label{fig:oracleexmaple}
\end{figure}

\begin{figure*}[h!]
    \noindent
    \centering
    \includegraphics[width=\linewidth]{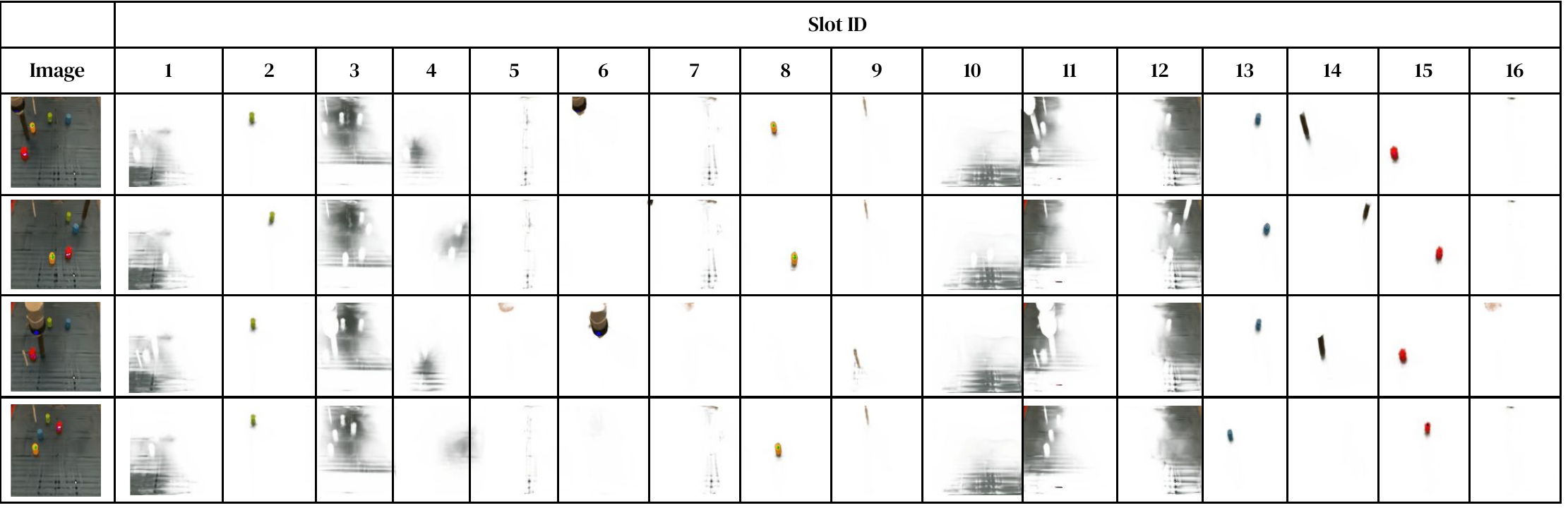}
    \caption{\textbf{Qualitative example of masks learned by Slot Attention on real-world data.} Slot Attention is able to discover the pixels corresponding to the blocks (slots 2,8,13,15), the robot arm (slot 6), the end-effector (slot 14), and the pole (slot 9) without any labels.}
    \label{fig:slotqualibcpaperdata}
    \vspace{-10pt}
\end{figure*}

\subsection{Multi-object Goal-conditioned Policy Learning: Task and Metrics}
\label{task:policylearning}
In this experiment, we compare different representation learning techniques by studying their effectiveness as inputs to a policy learning method. We only consider the imitation learning setup and learn a policy from a dataset of demonstrations provided by experts. The task for these experiments is to manipulate one of 8 blocks (i.e. the target block) on a plane to the target location shown by a purple rod. When the block is within 0.05 units of the rod, the episode is considered a success. If the robot fails to move the target block to the rod in 200 steps, the episode is considered a failure. In Figure \ref{fig:oracleexmaple}, we show an example of the demonstration of the task in the environment described in \ref{environment}. We use Implicit Behavior Cloning~\cite{florence2021implicit} to learn policies. During evaluation, we run the policy for 200 different initial configurations and measure the number of times the policy was able to successfully move the required block to the target location within the tolerated distance. We run policy training with 4 random seeds and report the mean and standard deviation of the success rates over the 4 runs. We present the results in Table~\ref{tab:policy_learning}. To compare different methods, we keep the policy learning method the same while we vary the input representations between RGB, RGB + Ground Truth (GT) Segmentation, Slot Attention, RGB+Slot Attention, Autoencoder, RGB+Autoencoder. To fairly compare between the different representation learning techniques we take the penultimate layer of the CNN encoder (before the spatial average pooling) and resize it to match the input RGB image. We optionally concatenate these features to the RGB image which is provided as input to the policy learning network. We also experimented with using MoCo features as input to IBC but were not able to train it to convergence. Remember that MoCo is trained to minimize a contrastive loss, and to succeed at minimizing this loss, the final representation does not need to capture information about all the objects in the scene. MoCo's loss can be minimized by focusing on objects that move more often than not, like the robot arm. The lack of convergence we observed for MoCo applied to IBC is likely due to MoCo features not reliably localizing the purple rod which is needed to solve the task. We refer the reader to \cite{florence2021implicit} for additional details on the IBC training.

\subsection{Multi-object Goal-conditioned Policy Learning: Results}

We make the following observations (Figure \ref{fig:ibc_lowdata}, Table \ref{tab:policy_learning}):

\begin{figure}
    \noindent
    \centering
    \includegraphics[width=\linewidth]{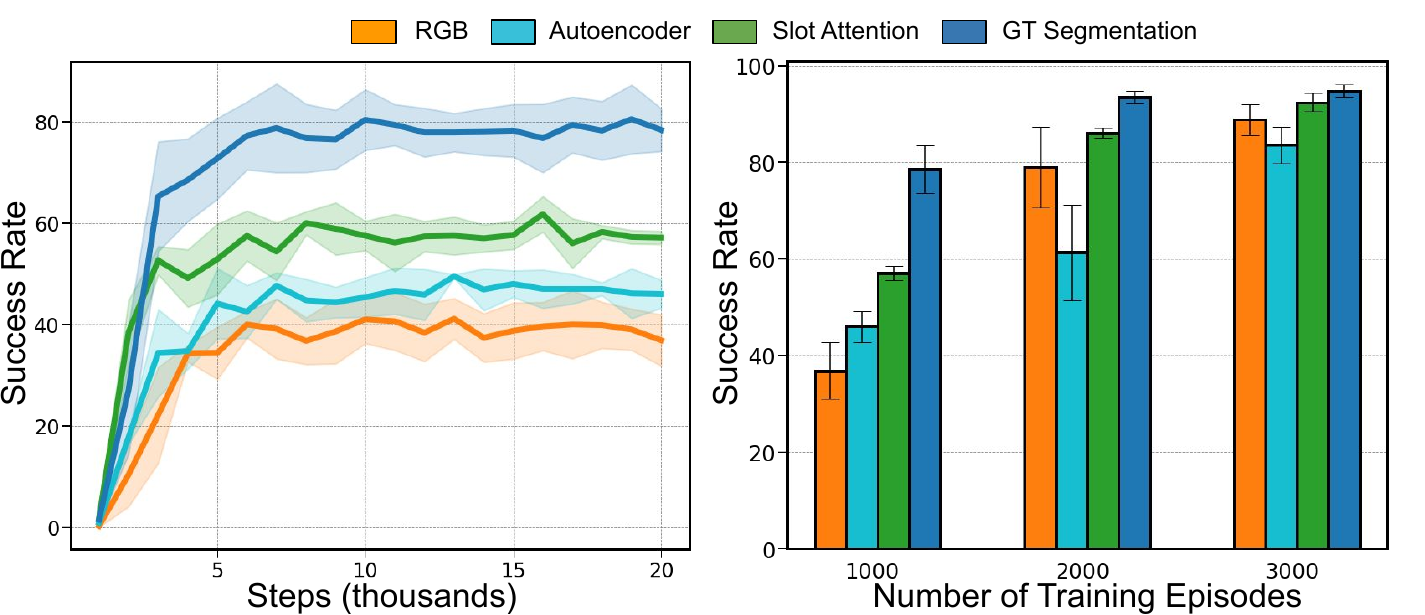}
    \caption{(Left) Validation performance of different methods during IBC training in low data regime (1000 episodes). It can be observed that using Slot Attention leads to a 20\% performance increase. As an upper bound, using ground truth segmentation masks resulted in about 40\% performance improvement. Solid lines show the mean across 4 seeds and the shaded area indicates 1 standard deviation from each side. (Right) Performance comparison on policy learning over episodes in training data. Slot Attention based representations provide a performance boost in the low data regimes.}
    \label{fig:ibc_lowdata}
    \vspace{-5pt}
\end{figure}

\noindent \textbf{Better perception inputs lead to more sample efficient policies.} As shown in Table \ref{tab:policy_learning}, by using the GT semantic segmentation for all the blocks as input, the policy learning method can learn high-performing policies with over 90 percent success rate with few samples (2000 episodes). However, the policy with RGB needs somewhere between 3000 and 10000 episodes to achieve the same performance. This motivates us to look for better input representations than raw RGB for IBC.
In the following experiments, we used a Slot Attention model trained with 16 bins since it had the lowest evaluation reconstruction loss during Slot Attention training. Both models were trained to convergence. 

\noindent \textbf{Slot Attention provides a performance boost in low data regimes.} We observe that Slot Attention models provide a boost in performance in the success rate of task completion over using raw RGB as input. Slot Attention models are object aware and by using this prior, we are able to learn representations from demonstration video datasets that can result in performance improvement without collecting object bounding boxes or segmentation masks from humans. We also note that the performance gain of using Slot Attention over baselines decreases as the number of samples available for learning the policy increases as shown in Figure \ref{fig:ibc_lowdata}.

% \noindent \textbf{Slot Attention is better than MoCo.} While MoCo has proven to be an effective representation learning technique for ImageNet datasets, for tasks where all object locations need to be encoded it might not be best. In section so and so we showed MoCo does not encode location of multiple objects. We hypothesize that is the reason it is not able to provide any performance boost when used as input to the policy learning network. We also tried only using MoCo features without RGB and the resulting experiments were unable to successfully execute the task. As a result only the performance of MoCo+RGB is reported in the table. 

%given that only having information about the end effector is %not sufficient for successful executation of the task

\noindent \textbf{Slot Attention performs better than Autoencoder.} We find that slot has better performance than autoencoder which has the same loss as the Slot Attention model but not the object/slot prior in its architecture. This shows that the prior of objects/slots is important for the performance gains.

\begin{figure}[h!]
    \noindent
    \centering
    \includegraphics[width=0.7\linewidth]{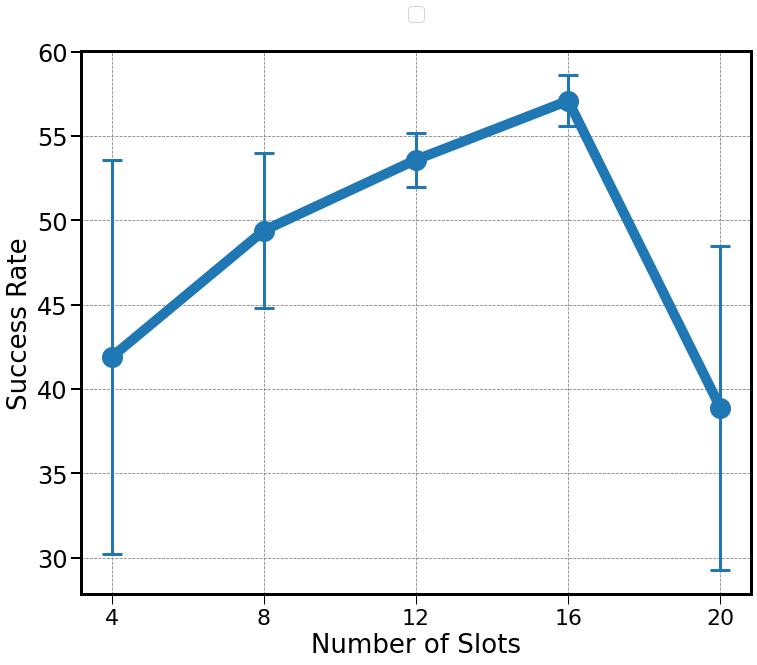}
    \caption{Effect of number of slots on IBC policy performance. Here we are using Slot Attention features in 1000 episodes data regime.}
    \label{fig:ibc_slotnumvar}
    \vspace{-10pt}
\end{figure}

\subsection{Ablation: effect of number of slot bins}
\label{ablation}
The number of slots is an important design choice in the Slot Attention architecture. This is a hyper-parameter that can be set by looking at the Slot reconstruction loss during the representation pretraining as well as evaluating on a validation set for downstream tasks. Figure \ref{fig:ibc_slotnumvar} shows the effect of varying the number of slots ($K$ in Section \ref{approach:slot}). In a scene of about 12 objects (8 blocks, 1 pole, 1 robot arm, 1 table, 1 background), it is reasonable to assume that at least 12 bins might be required to be able to consistently detect all the objects. However, we notice performance improvements till $K=16$. We also observe a drastic decrease at $K=20$ which is a known shortcoming~\cite{locatello2020objectcentric} of the original algorithm that it struggles with higher number of slots. 

\subsection{Action Prediction on a Real Robot}
\label{sec:action_pred}
To demonstrate our model's capability to learn representations that encode useful information for real-world policy learning, we show the performance of our learnt representations on the proxy task of action prediction. We hypothesize that representations that can better predict the conducted action at any frame in a demonstration video will perform better in policy learning. In this experiment, we learn representations from real-world robot interaction videos. Our robot is an arm attached to a table with a cylindrical end-effector that can push objects around on the table. We collect our data in a demonstration setup where the expert is asked to push one randomly chosen block to the randomly placed white pole. Figure \ref{fig:oracleexmaple} shows an example demonstration. We collected a total of 1080 demonstrations (880-200 train-test split) by randomizing the location of the pole and the target block for each episode. Based on the simulation experiments, we use 16 slots in this experiment.

The downstream task is to predict the action (the end-effector movement direction) given the current frame and a one-hot task embedding (representing the color of the target object) as input. We train a small regression network on top of these input embeddings to predict the action. We report the mean squared error between the predicted action from the small regression network and the expert action on the validation set as metric for evaluation. We present the results of this experiment in Table \ref{tab:action_pred} which shows that using Slot Attention provides a performance boost especially in the low-data regime. While only using Slot representations as input provide a performance boost over RGB, we observed that models with combined Slot representations and raw RGB as input perform best. For example, Slot+RGB combination decreases the prediction error by 13.6\% compared to RGB only models when training using 55 demonstrations (0.0625 fraction of the train dataset).

\begin{table}[]
\centering
\caption{Action prediction error (MSE) in real-world demonstrations.}
\begin{tabular}{l|c|c|c|c|c}
&   \multicolumn{5}{c}{\textbf{Fraction of Training Set}}\\
\midrule
\textbf{Input to action predictor} & \multicolumn{1}{c|}{\textbf{0.0625}} & \multicolumn{1}{c|}{\textbf{0.125}} & \multicolumn{1}{c|}{\textbf{0.25}}&  \multicolumn{1}{c|}{\textbf{0.5}} & \multicolumn{1}{c}{\textbf{1.0}} \\
\midrule
% \textbf{Input to object localizer ($\downarrow$)} &     \multicolumn{2}{c} {\textbf{PCK@0.1}}\\
% \toprule
RGB&2.35&2.01&1.78&1.57&\textbf{1.41}\\
Slot&2.07&1.93&1.67&1.55&1.46\\
% \midrule
Slot+RGB   &\textbf{2.03} &\textbf{1.87}&\textbf{1.64}&\textbf{1.54}&\textbf{1.41}\\
% \midrule

\end{tabular}

\label{tab:action_pred}
\end{table}

\section{Limitations}
Information about the approximate number of objects in the scene is needed to find the optimal number of slots. As observed in the original Slot Attention paper, too few or too many slots can result in degraded performance as it will not be able to properly segment the scene (too few) or will divide one object into multiple slots (too many). However, as shown here, it is possible to learn the optimal number of slots in simulation and use that information to learn slots on real data. 

%However, the reconstruction loss value during Slot training can be used as a general guide to choose this number.
% Similar to other self supervised methods, Slot Attention sometimes struggles when objects have the same color but a different shape. It will also confuse multiple instances of similar object leading to slot swapping. However, spatial information about the object location in the slot masks can potentially be used to differentiate these objects in cases where only a specific one of the multiple instances is the intended object. 

%Increasing the number of slots increases the time taken during training.

% Similar to the baselines, the algorithm will struggle with generalization to unseen objects with significantly different shape and color than those in training. However, the original Slot Attention paper showed evidence of the model generalizing to scenes with more objects than it was originally trained on at test time using the CLEVR6 dataset \cite{locatello2020objectcentric}.

% Slot Attention based models are able to push the performance of RGB-only models closer to the upper bound of models that have access to the ground truth segmentation masks. This suggests that self-supervised object-aware representations are a promising sample-efficient direction for augmenting visual input when learning policies.

\section{Conclusion and Future Work}
In this paper, we presented a method to improve performance of multi-object goal-conditioned behavior cloning policies using the Slot Attention architecture. We find that features and masks from this model are especially useful in the low data regime which is especially pertinent to deploying machine learning models on real-world robots. As preliminary evidence, we show a qualitative example of the Slot Attention algorithm successfully localizing objects in a real scene trained with no labels using demonstration data in Figure \ref{fig:slotqualibcpaperdata}. We also provide quantitative evidence of the usefulness of Slot Attention in the low data regime for the action prediction task in Section \ref{sec:action_pred}. It can be seen that even though the hole pattern on the table and the lighting complicates the task, Slot Attention is still able to discover all the objects in the scene successfully, including the thin rod. Although we study the utility of Slot attention in 2D pushing tasks, extensions of the Slot attention algorithm have shown promising results in segmenting 3D geometries in real-world robotic grasping videos \cite{kipf2022conditional} with minimal input conditioning signals. Studying the utility of these representations for complex 3D robotic tasks is an interesting future research direction. 

% ---- Bibliography ----

\bibliographystyle{IEEEtranN}
\newpage
\bibliography{root}
\end{document}